\documentclass[runningheads]{llncs}
\usepackage{multirow}
\usepackage{times}
\usepackage{epsfig}
\usepackage{amsmath}
\usepackage{amssymb}
\usepackage{booktabs}
\usepackage{color}
\usepackage[colorlinks]{hyperref}

\usepackage{graphicx}

\begin{document}
\title{You Only Learn Once: Universal Anatomical Landmark Detection}

\author{Heqin Zhu\inst{1} \and Qingsong Yao\inst{1} \and Li Xiao\inst{1} \and S.kevin Zhou\inst{1,2}}

\authorrunning{H. Zhu et al.}

\institute{\
Key Lab of Intelligent Information Processing of Chinese Academy of Sciences (CAS), Institute of Computing Technology, CAS, Beijing 100190, China\\
\email{s.kevin.zhou@gmail.com}  \and
Medical Imaging, Robotics, and Analytic Computing Laboratory and Engineering (MIRACLE) 
School of Biomedical Engineering \& Suzhou Institute for Advanced Research, University of Science and Technology of China, Suzhou 215123, China}

\maketitle

\begin{abstract}
Detecting anatomical landmarks in medical images plays an essential role in understanding the anatomy and planning automated processing. In recent years, a variety of deep neural network methods have been developed to detect landmarks automatically. However, all of those methods are unary in the sense that a highly specialized network is trained for a single task say associated with a particular anatomical region. In this work, for the first time, we investigate the idea of ``You Only Learn Once (YOLO)'' and develop a universal anatomical landmark detection model to realize multiple landmark detection tasks with end-to-end training based on mixed datasets. The model consists of a local network and a global network: The local network is built upon the idea of universal U-Net to learn multi-domain local features and the global network is a parallelly-duplicated sequential of dilated convolutions that extract global features to further disambiguate the landmark locations. It is worth mentioning that the new model design requires much fewer parameters than models with standard convolutions to train. We evaluate our YOLO model on three X-ray datasets of 1,588 images on the head, hand, and chest, collectively contributing 62 landmarks. The experimental results show that our proposed universal model behaves largely better than any previous models trained on multiple datasets. It even beats the performance of the model that is trained separately for every single dataset. Our code is available at \href{https://github.com/ICT-MIRACLE-lab/YOLO\_Universal\_Anatomical\_Landmark\_Detection}{https://github.com/ICT-MIRACLE-lab/YOLO\_Universal\_Anatomical\_Landmark\_Detection}.

\keywords{Landmark Detection \and Multi-domain Learning}
\end{abstract}

\section{Introduction}
Landmark detection plays an important role in varieties of medical image analysis tasks~\cite{zhou2019handbook,zhou2021review}. For instance, landmarks of vertebrae are helpful for surgery planning~\cite{chiras1997percutaneous}, which determines the positions of implants. Furthermore, the landmark locations can be used for segmentation~\cite{ref_seg} and registration~\cite{ref_reg} of medical images.

Because it is time-consuming and labor-intensive to annotate landmarks manually in medical images, many computer-assisted landmark detection methods are developed in the past years. These methods can be categorized into two types: traditional and deep learning based methods. Traditional methods aim at designing image filters and extracting invariant features, such as SIFT~\cite{ref_sift}. Liu et al.~\cite{liu2010search} present a submodular optimization framework to utilize the spatial relationships between landmarks for detecting them. Lindner et al.~\cite{ref_lindner} propose a landmark detection algorithm in the use of supervised random forest regression. These methods are less accurate and less robust in comparison to deep neural network methods. Yang et al.~\cite{ref_di2in} make use of a deep neural network and propose a deep image-to-image network built up with an encoder-decoder architecture for initializing vertebra locations, which are evolved with another ConvLSTM model and refined by a shape-based network. Payer et al.~\cite{ref_scn} propose a novel CNN-based neural network which integrates spatial configuration into the heatmap and demonstrate that, for landmark detection, local features are accurate but potentially ambiguous, while global features eliminate ambiguities but are less accurate~\cite{lay2013rapid,zhou2010shape}. Recently, Lian et al.~\cite{ref_mtn} develop a multi-task dynamic transformer network for bone segmentation and large-scale landmark localization with dental CBCT, which also makes use of global features when detecting landmarks.

However, all of those methods are unary in the sense
that a highly specialized network is trained for a single task say associated with
a particular anatomical region (such as head, hand, or spine), often based on a single dataset and not robust enough~\cite{qsyao2020landmarkattack}. 
It's promising and desirable to develop a model which is learned once and works for all the tasks\cite{ref_u2net,ref_li_uni}, that is ``You Only Learn Once''.
We, for the first time in the literature, develop a powerful, \textbf{universal model} for detecting the landmarks associated with different anatomies, each exemplified by a dataset. Our approach attempts to \textit{{unleash the potential of ``bigger data''}} as it utilizes the aggregate of all training images and builds a model that outperforms the models that are individually trained. We believe that \textit{{there are common knowledge among the seemingly different anatomical regions}}, observing that the local features of landmarks from different datasets share some characters (such as likely locating at corners, endpoints, extrema of curves or surfaces, etc.); after all, they are all landmarks. We attempt to design a model that is able to capture these common knowledge to gain more effectiveness while taking into account the differences among various tasks. To the best of our knowledge, this marks \textit{first such attempt} for landmark detection.

\begin{figure}[h]
        \includegraphics[width=\textwidth]{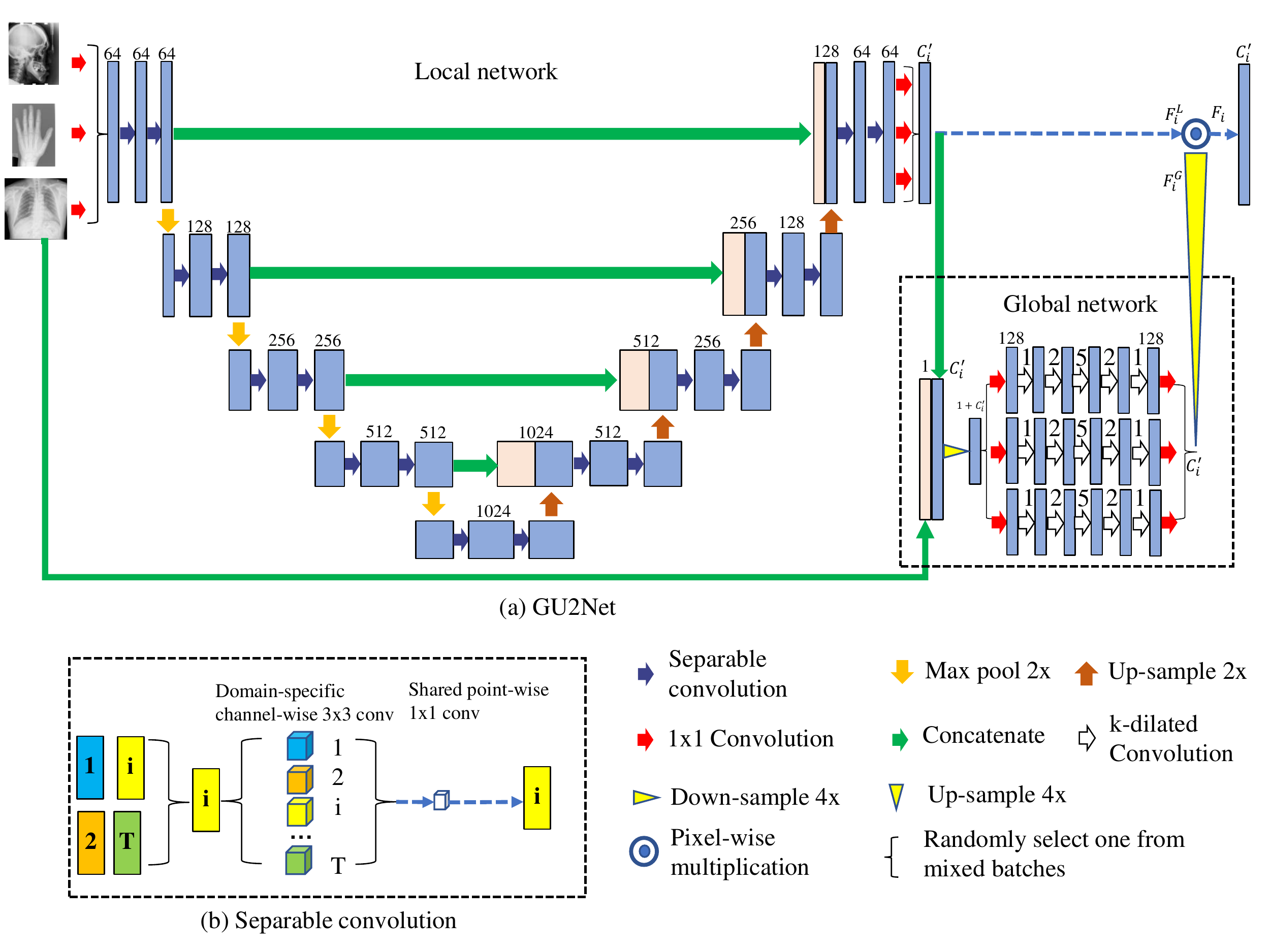}
        \caption{(a) The structure of the GU2Net model, consisting of two parts, namely the local network and the global network. The local network is a U-Net structure with each convolution replaced with separable convolution which consists of channel-wise convolution and point-wise convolution. The global network is parallel-duplicated sequential of five dilated small-kernel-size convolutions. (b) Separable convolution. Each 3x3 convolution is followed by batch-normalization and Leaky ReLU activation. The global network takes four times downsampled image and local heatmap as input and outputs four times upsampled heatmap.}
        \label{fig_overall}
\end{figure}

Our model, named as Global Universal U-Net (\textbf{GU2Net}), is inspired by the universal design of Huang et al.~\cite{ref_u2net} and the local-global design of Payer et al.~\cite{ref_scn}, reaping the benefits of both worlds. As shown in Figure~\ref{fig_overall}, it has a local network and a global network. Motivated by our observation, the local network is designed to be similar to a universal U-Net, which is a U-Net~\cite{ref_unet} with each convolution replaced with \textit{separable convolution}. The separable convolution is composed of \textit{channel-wise convolution} and \textit{point-wise convolution}~\cite{ref_u2net}, which model task-shared and task-specific knowledge, respectively, and hence has fewer parameters than a normal convolution. Differently, instead of outputting a segmentation mask in \cite{ref_u2net}, we output a landmark heatmap.
The local network extracts local features which are mostly accurate but still possibly ambiguous for landmarks. The global network is designed to extract global features and further reduce ambiguities when detecting landmarks. 
Different from Payer et al.~\cite{ref_scn}, our global network takes downsampled image and local features as input and uses dilated convolutions for enlarging receptive fields. 

{In sum, we make the following contributions:
\begin{itemize}
    \item The {\it first attempt} in the literature, to the best of our knowledge, to develop a universal landmark detection model that works for multiple datasets and different anatomical regions, unleashing the potential of ``bigger data'';
    \item {\it State-of-the-art performances} of detecting a total of 62 landmarks based on three X-ray datasets of head, hand, and chest, totaling 1,588 images but using only one model which needs fewer parameters.
\end{itemize}
}

\section{Method}

\textbf{Problem definition: }Let $\{D_1,D_2,\dots,D_T\}$ be a set of $T$ datasets, which are potentially from different anatomical regions. Giving an image $X_i\in R^{C_i\times H_i\times W_i}$ from Dataset $D_i$ along with corresponding landmarks $\{(x_{iC_i^{'}},y_{iC_i^{'}})\}$, we obtain the $k^{th}$ ($k \in [1,2,\ldots,C_i^{'}]$) landmark' heatmap $Y_{ik} \in R^{C_i^{'}\times H_i\times W_i}$ by using Gaussian function: 
\begin{equation}
    Y_{ik}= \frac{1}{\sqrt{2\pi}\sigma}\exp({-\frac{(x-x_{ik})^2+(y-y_{ik})^2}{2\sigma^2}}),
\label{Eq:gt}
\end{equation}
where $C_i$ is the number of channels on the input image (i.e. $C_i=1$ for a X-ray image); $C_i^{'}$ is the number of channels on output heatmap, namely the number of landmarks; $H_i$ is the height of image $X_i$ and $W_i$ is the width of image $X_i$. 

\subsection{The Local Network} \label{sec.1}
The local network $\phi_{LN}$ aims at extracting local features and generating a local heatmap that is used to determine the accurate location of a landmark. 
To work well on all datasets, we replace the standard convolution in U-Net with separable convolution which consists of domain-specific channel-wise convolution and shared point-wise convolution. Each data set is assigned a different channel-wise convolution separately, while all datasets share the same point-wise convolution. 
It extracts local feature from input $X_i$, following a local heatmap $\tilde{F}_i^L$:
\begin{equation}
    \tilde{F}_i^L= \phi_{LN}(X_i;\theta_{di}^L,\theta_s^L)
\end{equation}
where $\theta_{di}^L$ is the parameter of domain-specific channel-wise convolution corresponding to $D_i$; and $\theta_s^L$ is the parameter of shared point-wise convolution. In separable convolution, considering a N-channel input feature map and a M-channel output feature map,  we firstly apply $N$ channel-wise filters in the shape of $R^{3 \times 3}$ to each channel and concatenate the $N$ output feature maps. Secondly, we apply $M$ point-wise filters in shape of $R^{1\times 1\times N}$ to output the feature maps of $M$ channels~\cite{ref_u2net}. Accordingly, the total number of parameters is $9\times N\times T+N\times M$, while it's $9\times N\times M\times T$ for $T$ standard $3\times 3$ convolutions.

\subsection{The Global Network} \label{sec.2}
Global structural information plays an essential role in landmark detection~\cite{ref_scn,ref_mtn}, which motivates us to design an additional global network $\phi_{GN}$. $\phi_{GN}$ is composed of five dilated $3\times 3$ convolutions with dilations being $[1,2,5,2,1]$. With the enhancement of dilated convolution, $\phi_{GN}$ achieves large receptive field, which is benefit for capturing the important global information~\cite{ref_dilation}. Since different anatomical regions vary a lot in appearance, we duplicate our global network for each dataset (see Figure.~\ref{fig_overall}, resulting in domain-specific parameters $\theta _{di}^G$. As shown in Figure~\ref{fig_overall}(a)), $\phi_{GN}$ takes image $X_i$ and local feature $\tilde{F}_i^L$ as input and aggregates the global information at a coarse-grained scale, flowing global heatmap $\tilde{F}_i^G$:
\begin{equation}
    \tilde{F}_i^G= \phi_{GN}(X_i,\tilde{F}_i^L;\theta_{di}^G)
\end{equation}

\subsection{Loss Function} \label{sec.3}
As shown in Figure~\ref{fig_overall}(a), we combine the local and global information by point-wise multiplying the local heatmap $\tilde{F}_i^L$ and global heatmap $\tilde{F}_i^G$, resulting in final heatmap $\tilde{F}_i = \tilde{F}_i^G \odot \tilde{F}_i^L$, where $\odot$ is the pixel-wise multiplication.
In the training stage, we penalize the final heatmap $\tilde{F}_i$ and the ground truth $Y_{i}$ (defined in Eq.~\ref{Eq:gt}):
\begin{equation}
L_i = \sum_{y\in Y_i, f\in \tilde{F}_i}-y\log{f}-(1-y)\log{(1-f)}
\end {equation}
In the inference stage, the k-th landmark is obtained after finding maximum location of the k-th channel in output heatmaps $\tilde{F_i}$. 
\begin{equation}
    \text{landmark}_k = arg max(\tilde{F}_{ik})
\end{equation}

\section{Experiments}
In this section, we qualitatively and quantitatively evaluate our universal model and compare it with the state-of-the-art methods on three public X-ray datasets of head, hand, and chest. Except as otherwise noted, models are learned only once on the three datasets which are randomly mixed by batch. Evaluation is carried out on a single dataset separably for each model. Furthermore, we conduct an ablation study to demonstrate how different components help to improve the performance of our universal model.

\subsection{Settings}
Our deep networks are implemented in Pytorch 1.3.0 and run on a TITAN RTX GPU with CUDA version being 11. Each $3\times 3 $ convolution is followed by batch normalization~\cite{ref_batnorm} and Leaky ReLU activation~\cite{ref_relu}. We convert landmarks to a Gaussian heatmap which retains the probablility distribution of landmark in each pixel, with $\sigma$ set to 3. 

 For data augmentation, we rotate the input image by 2 degrees with 0.1 probability and do the translation by 10 pixels in each direction with 0.1 probability. When training networks, we set batch-size to 4 and learning-rate to [1e-4,1e-2]. The binary cross-entropy (BCE) loss and an Adam optimizer are used to train the network up to 100 epochs and a cyclic scheduler~\cite{ref_clr} is used to decrease learning rate from 1e-2 to 1e-4 dynamically. For evaluation, the inference model is chosen as the one with minimum validation loss and evaluated on two metrics: mean radial error (MRE) and successful detection rates (SDR).
\subsection{Dataset}
\subsubsection{Head} \label{sec_head}
The head dataset is an open-source dataset that contains 400 cephalometric X-ray images~\cite{ref_head}. We choose the first 150 images for training and the other 250 images for testing. Each image is of size $2400\times 1935$ with a resolution of $0.1mm \times 0.1mm$. When training networks on the head dataset, we resize the original image to the size of $512\times 416$ to keep the ratio of width and height. There are 19 landmarks manually labeled by two medical experts and we use the average labels like \cite{ref_scn}. 

\subsubsection{Hand}
The hand dataset is also a public dataset\footnote{\href{https://ipilab.usc.edu/research/baaweb}{https://ipilab.usc.edu/research/baaweb}} which contains 909 X-ray images. The first 609 images are used for training and the other 300 images are used for testing. The sizes of these images are not all the same, so we resize images to the shape of $512\times 368$. To calculate the physical distance and compare it with other methods, we assume that the width of the wrist is 50mm, following Payer et al.~\cite{ref_scn}. With the two endpoints of the wrist being $p,q$, the physical distance can be calculated as the multiplication of the pixel distance and $\frac{50}{\|p-q\|_2}$. A total of 37 landmarks have been manually labeled by Payer et al.~\cite{ref_scn} and the two endpoints of the wrist can be obtained from the first and fifth points, respectively.

\subsubsection{Chest}
We also adopt a public chest dataset\footnote{\href{https://www.kaggle.com/nikhilpandey360/chest-xray-masks-and-labels}{https://www.kaggle.com/nikhilpandey360/chest-xray-masks-and-labels}} which contains two subset: the China set and the Montgomery set~\cite{ref_chest1,ref_chest2}. We select the China set and exclude cases that are labeled as abnormal lungs(diseased lungs) to form our experimental dataset. Our chest dataset has 279 X-ray images. The first 229 images are used for training and the last 50 images are used for testing. Since the chest dataset has no landmark labels, we manually annotate six landmarks in each image. The left three landmarks are the top, the bottom, and the right boundary point of the right lung. It's the same with the right three landmarks. In the rough, the six landmarks determine the boundary of the lung (see Figure~\ref{fig_visual}). We resize the input image to the shape of $512\times 512$. Besides, since the chest dataset only contains png images and the physical spacing is not known, we use pixel distance to measure the model's performance.

\subsection{Evaluation}
Table~\ref{tab_results} shows the experimental results in comparison of different methods on the head, hand, and chest datasets, with different SDR thresholds. Images are resized during training and testing.

\begin{table}[!ht]
\centering
\caption{Quality metrics of different models on head, hand, and chest datasets. * represents the performances copied from the original paper. + represents the model is learned on the mixed three datasets. - represents that no experimental results can be found in the original paper. The best results are in \textbf{bold} and the second best results are \underline{underlined}. }
\label{tab_results}
\resizebox{0.9\linewidth}{!}{
\begin{tabular}{lrrrrrrrrrrrrr}
\hline
\multirow{2}*{Models}  &\multirow{2}*{MRE} &\multicolumn{4}{c}{Head  SDR(\%)}   &\multirow{2}*{MRE} & \multicolumn{3}{c}{Hand SDR(\%)} & \multirow{2}*{MRE} &  \multicolumn{3}{c}{Chest SDR(\%)}\\
\cline{3-6}\cline{8-10}\cline{8-10}\cline{12-14}   &(mm)& 2mm   & 2.5mm & 3mm   & 4mm   &(mm)     & 2mm & 4mm & 10mm  &(px)  & 3px & 6px & 9px\\
\hline
\multirow{1}*{Ibragimov et al.~\cite{ref_ibragimov}*}  &1.84& 68.13 & 74.63 & 79.77 & 86.87  &-    &-&-&-                  &-&-&-&-\\
\multirow{1}*{{\v{S}}tern et al.~\cite{ref_stern}*}    &-&-&-&-&-                         &\underline{0.80} & 92.20 & 98.45 & 99.83 &-&-&-&-\\
\multirow{1}*{Lindner et al.~\cite{ref_lindner}*}      &\underline{1.67}& 70.65 & 76.93 & 82.17 & \underline{89.85} &0.85 & 93.68 & 98.95 & 99.94 &-&-&-&-\\
\multirow{1}*{Urschler et al.~\cite{ref_urschler}*}    &-& 70.21 & 76.95 & 82.08 & 89.01     &\underline{0.80}  & 92.19 & 98.46 & \underline{99.95} &-&-&-&-\\
\multirow{1}*{Payer et al.~\cite{ref_scn}*}            &-& \underline{73.33} & \underline{78.76} & \underline{83.24} & 89.75   &\textbf{0.66} & \underline{94.99} & \underline{99.27} & \textbf{99.99} &-&-&-&-\\
\multirow{1}*{U-Net~\cite{ref_unet}+}                  &12.45& 52.08 & 60.04 & 66.54 & 73.68    &6.14  & 81.16 & 92.46 & 93.76 &\underline{5.61} &\underline{51.67} & \underline{82.33} & \textbf{90.67}\\ 
\multirow{1}*{GU2Net (Ours) + }                        &\textbf{1.54}& \textbf{77.79} & \textbf{84.65} & \textbf{89.41} & \textbf{94.93}   &0.84& \textbf{95.40} & \textbf{99.35} & {99.75} &\textbf{5.57}& \textbf{57.33} &\textbf{82.67} & \underline{89.33}\\
\hline
\end{tabular}

}
\end{table}

On the head dataset, our GU2Net achieves the best accuracy within all thresholds (2mm, 2.5mm, 3mm, 4mm) and obtains a MRE of $1.54\pm 2.37$ mm, behaving much better than U-Net which is also learned on the mixed multiple datasets. GU2Net even beats all models marked with *  which are learned on a single dataset. Within 2mm, GU2Net achieves the best SDR of 77.79\%, outperforming the previous state-of-the-art method ~\cite{ref_scn} by 4.46\%. Such an improvement is consistent among SDRs at other distances.

On the hand dataset, our GU2Net also reaches the best accuracy of 95.40\% within 2mm and 99.35\% within 4mm which is far ahead of other models learned on the single hand dataset. GU2Net also performs better than U-Net which is learned on the mixed multiple datasets by a margin of 14.24\% within 2mm. For SDR within 10mm, the performance of GU2Net is not the best but very close to the best.

On the chest dataset, our GU2Net obtains an MRE of $5.57\pm 20.54$ px, behaving better than U-Net within 3px and 6px, but a little worse than U-Net within 9px. This is probably due to that for the head and hand datasets, the landmarks are mostly with bone structures, while here the landmarks are with the lung, which concerns soft-tissue contrast. This motivates us in the future to design a new universal model to better capture the nuances between bone and soft tissue. 

In summary, our proposed GU2Net generally outperforms any other state-of-the-art methods learned on a single dataset or mixed multiple datasets, especially under high-precision conditions, which is evident from the SDR values within say 2mm, 4mm, 3px, and 6px. 

\subsection{Ablation Study}
In order to demonstrate the effectiveness of our local network $\phi_{LN}$ and global network $\phi_{GN}$, we perform ablation study on the mixed dataset by merging the three datasets together. There are total $250+300+50=600$ images for testing. The average MRE and SDR on the mixed dataset are adopted as metrics. We evaluate the performance on U-Net, Tri-UNet, $\phi_{GN}$, $\phi_{LN}$(with local network only) and $\phi_{GN}$(with global network only), and GU2net. Here Tri-UNet is a U-Net with each convolution duplicated 3 times, and with ReLU replaced with leaky ReLU.

\begin{table}[!ht]
\centering
\caption{Ablation study of our universal model with local network and global network.$\theta_d$ denotes domain-specific parameters while $\theta_s$ denotes shared parameters.}
\label{tab_ablation}
\begin{tabular}{lrrrrrr}
\hline
\multirow{2}*{Models}    &\multicolumn{2}{c}{Parameter} & \multicolumn{1}{c}{MRE$\pm $STD}& \multicolumn{3}{c}{SDR(\%)}\\
\cline{2-4}\cline{5-7}            & number  & type                 & (px)                     & 2px   & 4px   & 6px   \\
\hline
\multirow{1}*{U-Net~\cite{ref_unet}}&17.3M   &$\theta_s$           & 7.99$\pm$26.14             & 71.67 & 86.40 & 89.26 \\
\multirow{1}*{Tri-UNet}             &51.92M  &$\theta_d$           & \underline{1.16}$\pm$5.21  & 88.19 & \underline{97.46} & \underline{99.14}\\
\multirow{1}*{$\phi_{GN}$}          &1.35M   &$\theta_d$           & 2.31$\pm$5.64  & 67.39 & 94.22 & 97.62\\
\multirow{1}*{$\phi_{LN}$}          &2.12M   &$\theta_d,\theta_s$  & 1.20$\pm$\underline{4.98}              & \underline{88.68} & 97.29 & 99.12\\
\multirow{1}*{GU2Net($\phi_{LN}\odot \phi_{GN}$)} &4.44M &$\theta_d,\theta_s$ & \textbf{1.14}$\pm$\textbf{3.94}& \textbf{89.09} & \textbf{97.49} & \textbf{99.52}\\
\hline
\end{tabular}
\end{table}

As shown in Table~\ref{tab_ablation}, when comparing $\phi_{LN}$ with U-Net, it is evident that separable convolution in the local network improves the model's performance by a large margin. Thus, the architecture of local network is more capable for multi-domain learning. By comparing the results of GU2Net with $\phi_{LN}$ and $\phi_{GN}$, we observe much improvement of detection accuracy, which demonstrates the effectiveness of fusing local information and global information. Thus global information and local information are equally important for accurate localization of anatomical landmarks. Since U-Net only has shared parameters, its performance is the worst among all models and falls behind others by a huge gap. U-Net is even inferior to our global network $\phi_{GN}$ which consists of small amount of domain-specific parameters. By comparing GU2Net with Tri-UNet and U-Net, our GU2Net behaves better than them under all the measurements even though the number of parameters of GU2Net is about 3 times less than that of U-Net and 10 times less than Tri-UNet's, which demonstrates the superiority of our architecture and the indispensability of both shared and domain-specific parameters.

\begin{figure*}[!t]
    \centering 
    \begin{minipage}[t]{0.19\textwidth}
        \centering
        \includegraphics[width=1\textwidth]{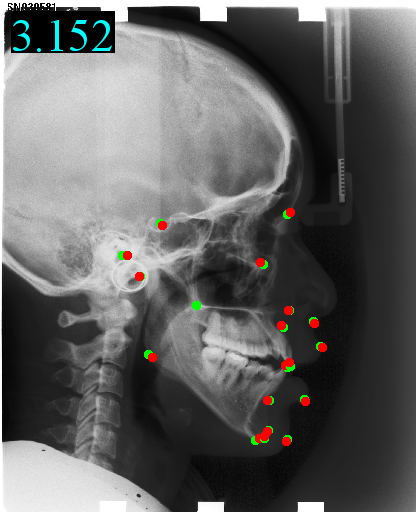}
        \includegraphics[width=1\textwidth]{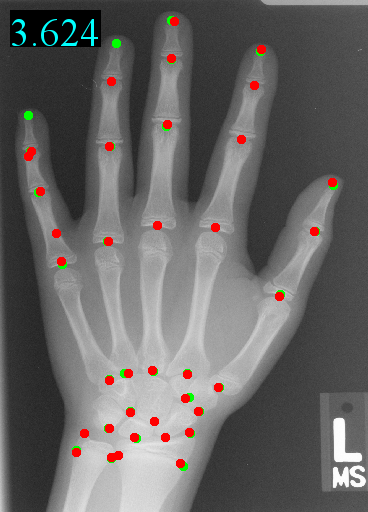}
        \includegraphics[width=1\textwidth]{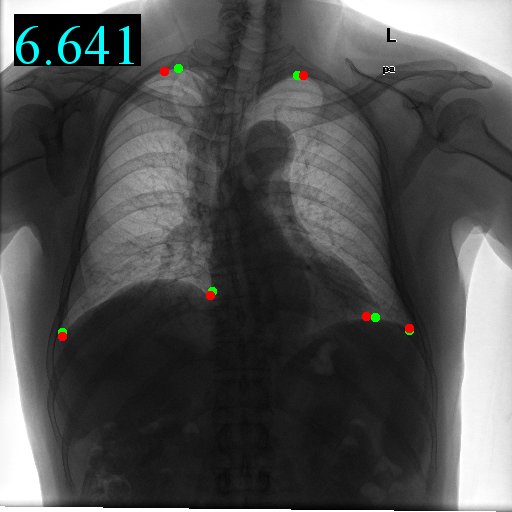}
        U-Net
    \end{minipage}
        \begin{minipage}[t]{0.19\textwidth}
        \centering
        \includegraphics[width=1\textwidth]{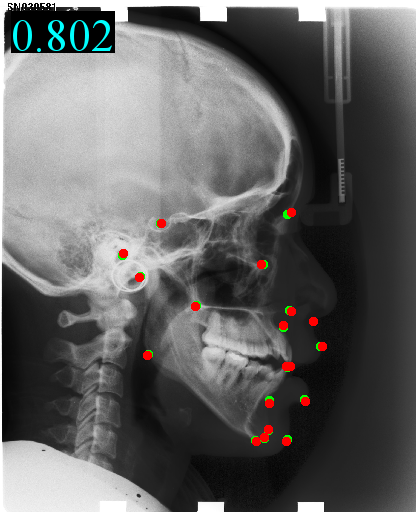}
        \includegraphics[width=1\textwidth]{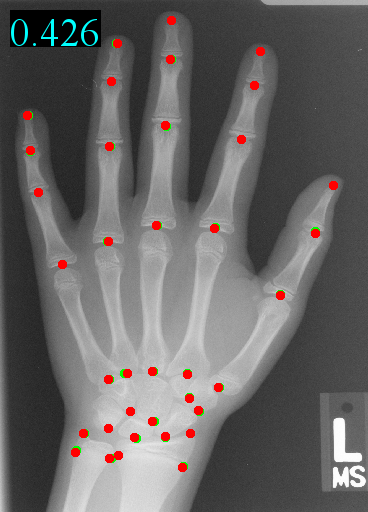}
        \includegraphics[width=1\textwidth]{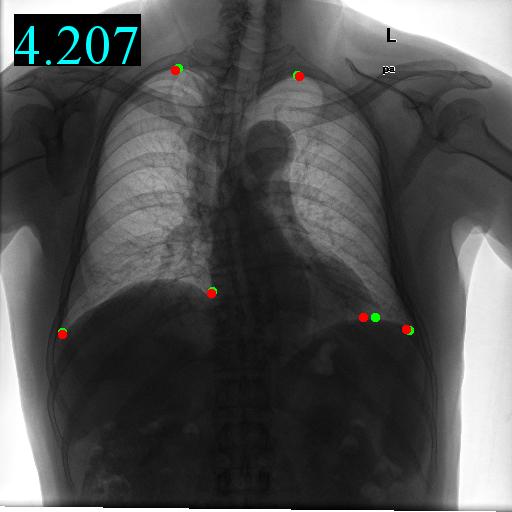}
        Tri-UNet
    \end{minipage}
        \begin{minipage}[t]{0.19\textwidth}
        \centering
        \includegraphics[width=1\textwidth]{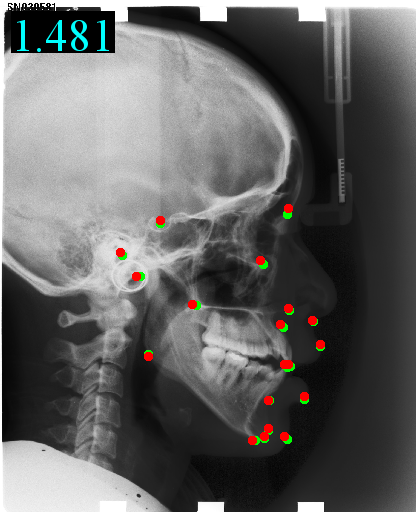}
        \includegraphics[width=1\textwidth]{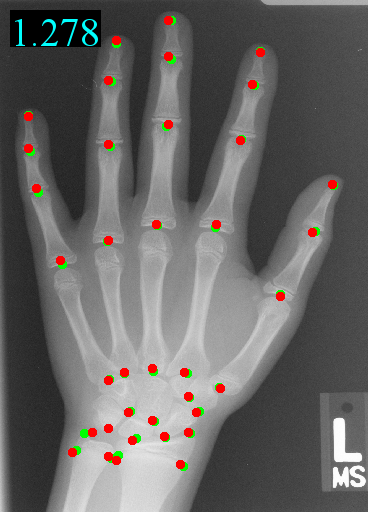}
        \includegraphics[width=1\textwidth]{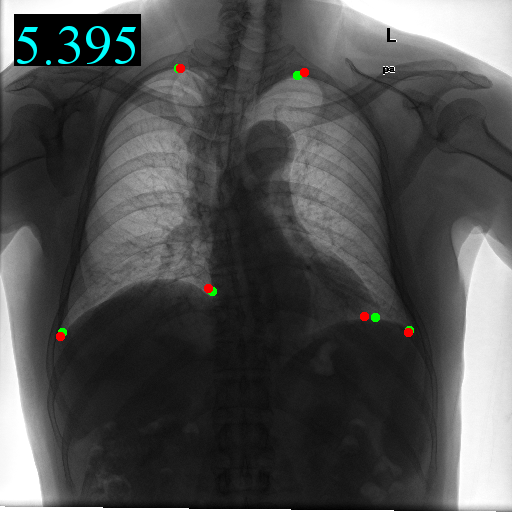}
        $\phi_{GN}$
    \end{minipage}
    \begin{minipage}[t]{0.19\textwidth}
        \centering
        \includegraphics[width=1\textwidth]{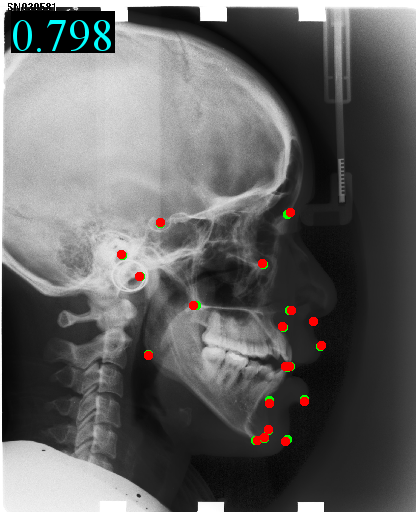}
        \includegraphics[width=1\textwidth]{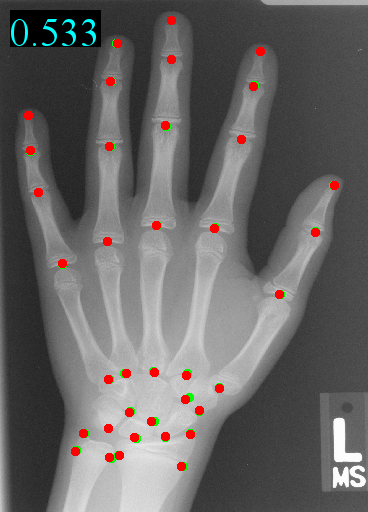}
        \includegraphics[width=1\textwidth]{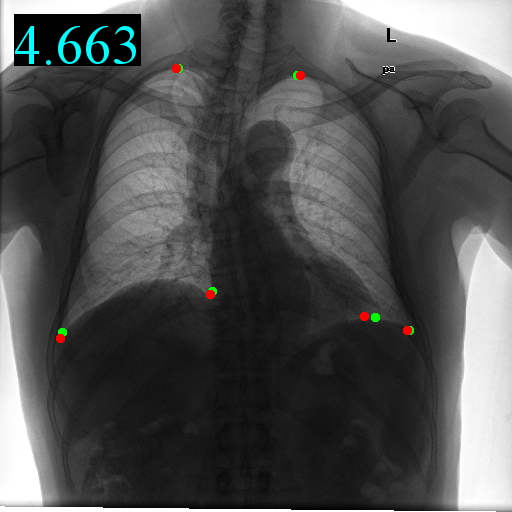}
        $\phi_{LN}$
    \end{minipage}
    \begin{minipage}[t]{0.19\textwidth}
        \centering
        \includegraphics[width=1\textwidth]{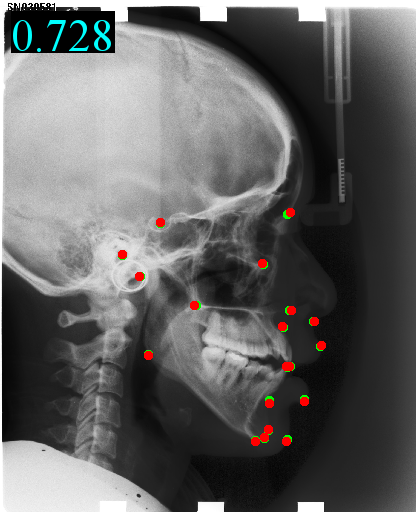}
        \includegraphics[width=1\textwidth]{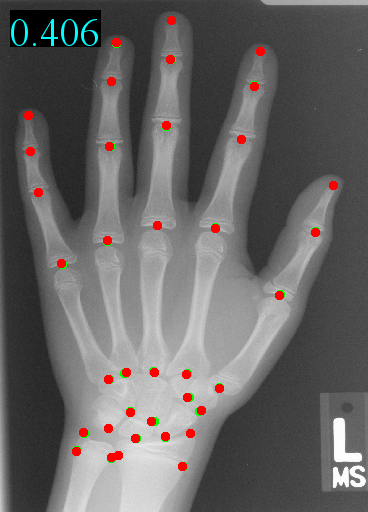}
        \includegraphics[width=1\textwidth]{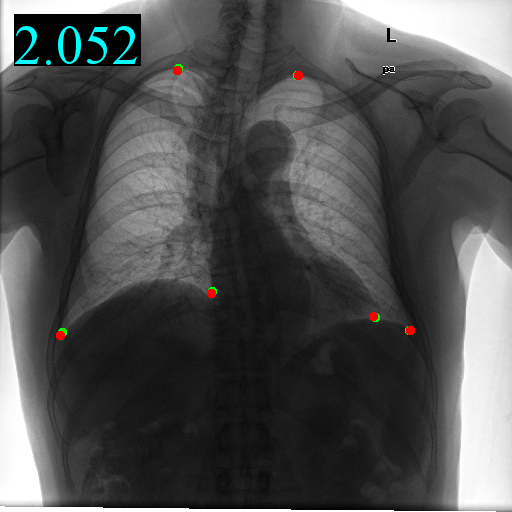}
        GU2Net
    \end{minipage}
    \caption{Qualitative comparison of different models on head, hand, and chest datasets. All images are randomly selected. The red points \textcolor{red}{$\bullet$} are the learned landmarks while the green points \textcolor{green}{$\bullet$} are the ground truth labels. The MRE value is displayed on the top left corner of the image for reference. }
    \label{fig_visual}
\end{figure*}

To qualitatively show the superiority of our universal model, we further visualize the learned landmarks in Figure~\ref{fig_visual}. The MRE value is displayed on the top left of the image for reference. The red points are the learned landmarks while the green points are the ground truth labels. It's clear that our model's results have more overlap regions of red and green points than other models', which also can be verified according to the MRE values.

\section{Conclusions}
{To build a universal model, we propose our global universal U-Net to integrate the global and local features, both of which are beneficial for accurately localizing landmarks. Our universal model makes the first attempt in the literature to train a single network on multiple datasets for landmark detection. Using the separable convolution makes it possible to learn multi-domain information, even with a reduced number of parameters. Experimental results qualitatively and quantitatively show that our proposed model performs better than other models trained on multiple datasets and even better than models  trained on a single dataset separately. Future works include designing different network architectures and exploring more datasets to further improve the performances on all datasets simultaneously.}

\clearpage
\bibliographystyle{splncs04}
\bibliography{paper}
\end{document}